\documentclass[letterpaper, 10 pt, conference]{ieeeconf}  

\IEEEoverridecommandlockouts                              

\usepackage{graphics} 
\usepackage{epsfig} 
\usepackage{mathptmx} 
\usepackage{times} 
\usepackage{amsmath} 
\usepackage{amssymb}  
\usepackage{color}

\usepackage{fancyhdr}

\definecolor{red}{RGB}{255,0,0}
\DeclareMathOperator*{\argmax}{\arg\!\max}

\title{\LARGE \bf
Multi-Type Activity Recognition in Robot-Centric Scenarios
}

\author{Ilaria Gori$^{1}$, J. K. Aggarwal$^{1}$, Larry Matthies$^2$, and M. S. Ryoo$^{3}$
\thanks{$^{1}$Ilaria Gori and J. K. Aggarwal are with the Department of Electrical and Computer Engineering,
        University of Texas at Austin, 1 University Station C0803, Austin, TX 78712, USA}%
\thanks{$^{2}$Larry Matthies is with Jet Propulsion Laboratory, California Institute of Technology, 4800 Oak Grove Drive, Pasadena, CA 91109, USA}%
\thanks{$^{3}$M. S. Ryoo is with the School of Informatics and Computing, Indiana University, 919 E. 10th Street, Bloomington, IN 47408, USA~%
        {\tt\small mryoo@indiana.edu}}%
}

\begin{document}

\maketitle
\thispagestyle{empty}
\pagestyle{empty}

\thispagestyle{fancy}
\fancyhf{}
\fancyhead[L]{The final version of this paper was published by the IEEE Robotics and Automation Letters (RA-L) 2016.}

\begin{abstract}
Activity recognition is very useful in scenarios where robots interact with, monitor or assist humans. In the past years many \emph{types} of activities -- single actions, two persons interactions or ego-centric activities, to name a few -- have been analyzed. Whereas traditional methods treat such types of activities separately, an autonomous robot should be able to detect and recognize multiple types of activities to effectively fulfill its tasks. We propose a method that is intrinsically able to detect and recognize activities of different types that happen in sequence or concurrently. We present a new \emph{unified} descriptor, called Relation History Image (RHI), which can be extracted from all the activity types we are interested in. We then formulate an optimization procedure to detect and recognize activities of different types. We apply our approach to a new dataset recorded from a robot-centric perspective and systematically evaluate its quality compared to multiple baselines. Finally, we show the efficacy of the RHI descriptor on publicly available datasets performing extensive comparisons.
\end{abstract}

\section{INTRODUCTION}
The recognition of activities is a crucial problem in computer vision. Usually, activities are categorized based on how many people participate (one person, two persons, a group of people), and the point of view from which they are recorded (third person, ego-centric). The combination of these two characteristics generates $6$ possible \textit{types} of activities. Traditional computer vision methods address the recognition of only one \textit{type} of activity. For example, single person activities from a third-person perspective \cite{laptev03} has been studied broadly in the past. Two-person interactions from a third person perspective \cite{ryoo09} and group activities from a third point of view \cite{khamis12} have been analyzed more recently. In the last few years, the ego-centric perspective has been largely explored in the form of ego-centric activities \cite{grauman13,fathi11}, where the goal is understanding what the person wearing the camera is doing, and ego-centric interactions \cite{ryoo13,lu15}, where a robot usually wears a camera, and the objective is classifying what activities other people are doing to the robot.

In all the above-mentioned cases, videos to be classified belong to a specific \textit{type} of activities. However, imagine a service robot, who provides directions and recommendations in public places, or a robot assistant, who helps elderly people, or yet a surveillance robot, who monitors crowded areas. In these situations, people usually perform several different types of activities at the same time, and the robot should be able to classify them all. For example, it should be able to understand if a person intends to talk to it (ego-centric interactions), to detect if someone needs help (single person activities from a third person perspective), or notify someone if there is a fight (two-persons interactions from a third person perspective). In summary, there is a clear need to move towards a more general objective: classifying more complex scenes where multiple people perform activities of different types concurrently or in a sequence. To our knowledge, ours is the first recognition work that considers such multiple types of activities. We provide a new video dataset recorded from a robot-perspective where multiple activities and interactions happen at the same time, and present a novel approach to appropriately analyze such videos.

\begin{figure*}
\begin{center}
\includegraphics[width=0.85 \textwidth]{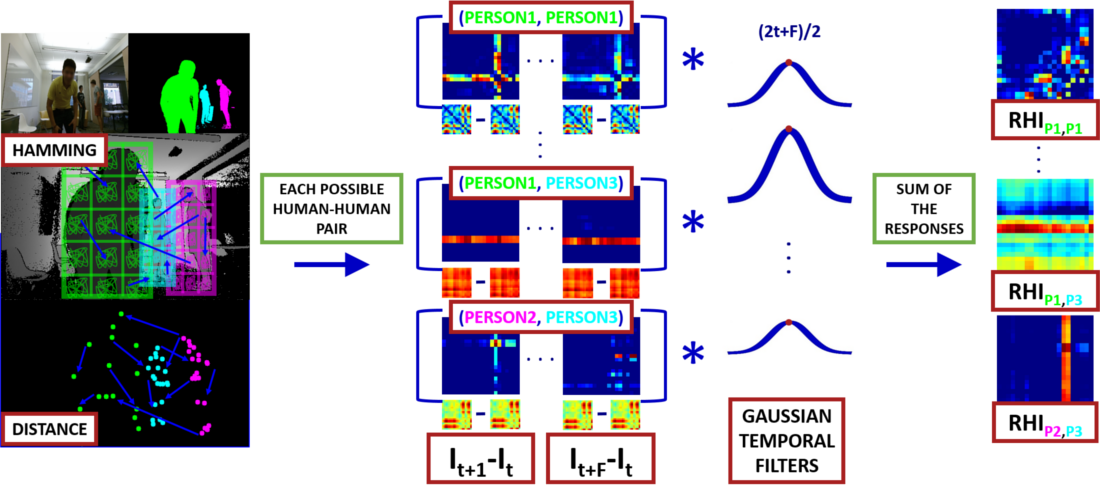}
\end{center}
\caption{This picture represents how RHIs on skeleton and depth are extracted from a single person (e.g., person1) and pairs of persons (e.g., $\{$person1, person3$\}$ and $\{$person2, person3$\}$). For RHIs on skeletons, the relations between pairs of joints are computed for each frame, for each pair. For RHIs on depth, the Hamming distance between strings obtained from the $\tau$ test represents the relations between local regions. The first descriptor in a temporal window is subtracted to each other frame in the window; these differences are then convoluted with a set of Gaussian filters. The filter responses are summed to generate the final RHI descriptor over the window of frames.}
\label{fig:rhiextr}
\end{figure*}
In order to discriminate different types of activities, a \textit{unified} descriptor, i.e., a descriptor that can be extracted from several activity types, is necessary. We propose a new \textit{unified} mid-level descriptor, called \textit{Relation History Image} (RHI) (see Fig. \ref{fig:rhiextr}). The Relation History Image is built as the variation over time of relational information between every pair of local regions (joints or image patches) belonging to one or a pair of subjects. It can be extracted from several types of activities while maintaining the same format and dimensionality, thereby allowing the direct comparison between activity videos of completely different types. This direct comparison enables the robot to detect ongoing activities and discard false positives even in the scenario where multiple types of activities are present in the same video. In contrast to the other previous appearance-based unified descriptors such as STIP \cite{laptev03} or DSTIP \cite{lu13}, RHI is suitable to represent subtle interactions where human body configurations are very important. The presented experiments confirm that RHI significantly outperforms state-of-the-art descriptors in our scenario, as well as on other public datasets.




The main peculiarity of multi-type activity videos is that it is impossible to know a priori who is performing what types of activities. In traditional videos with single person actions, it is assumed that each person is doing something singularly. In traditional videos of two-person interactions, it is expected that there are two persons and that they are interacting. In our case, if there are three persons in the scene, there is no way of knowing a priori if two of those persons are interacting and who they are, or if some of them are interacting with the robot, or if all three of them are on their own. To solve this problem, we propose a new method based on optimization. First, RHI descriptors are extracted from all the possible combinations of subjects in the scene, including those formed by pairing each subject with himself. All the combinations are fed to a classifier, which provides for each descriptor the confidence that it belongs to a certain class. Since RHI can be extracted from different types of activities, the resultant class can be an interaction, or an action, or an interaction with the robot. An optimization problem based on the generated confidence returns who is performing a certain type of activity, and what activity that is. In the example above, a possible outcome of our algorithm is: \textit{person1-person3 hug, person2 sit}. Notably, the optimization problem outputs both the action and who is involved in it.

Our main contribution is the idea of recognizing concurrent activities belonging to different \textit{types}, which is crucial in real-world scenarios. Our technical contributions are: 1) a new \textit{unified} descriptor, RHI, which can be extracted from multiple types of activities and thus simplifies the classification step; 2) a new method to identify what persons are involved in what activity; 3) a systematic evaluation to compare the proposed approach to several baselines on a newly collected dataset as well as on public ones.

\section{RELATED WORK}
\label{sec:related}
Most of the approaches in the literature  tackle the problems of third-person activity recognition \cite{wang11,wang12,kantorov2014,junejo08}, two persons interactions \cite{ryoo09,kong14,raptis13,li13} and group activities \cite{khamis12,mori10,ryoo11}. Li et al.~\cite{li13} for instance, build a tensor which contains relations between two persons' low-level features over different pairs of frames. In \cite{wang12}, joint relations and Local Occupancy Patterns (LOP) are used to retrieve a frame-based descriptor, then the Fourier Transform is used to model their temporal dynamics. Wang et al.~\cite{wang11} propose an effective method to create dense trajectories from which local features are extracted. In the last few years, also first-person activities \cite{fathi11,kanade12,grauman13} and first-person interactions \cite{ryoo13,lu15} have been analyzed. However, previous works are specific to only one type of activities. In contrast, our method explicitly addresses situations where the goal is recognizing different types of activities performed concurrently.

Detecting who is interacting with whom has been recently studied by Fathi et al.~\cite{fathi12}: given a set of persons, their goal is to classify monologues, dialogues, and discussions. Their method is based on detecting the social attention of the participants. Nonetheless, it is assumed that only one social interaction happens in each scene. Our goal is more general, as we consider cases where there are multiple social/physical activities performed at the same time.

In a different category, Lan et al.~\cite{lan12} analyze videos with multiple people and formulate a hierarchical classifier to recognize single activities at a low level, while inferring their social roles at an intermediate level. However, they only focused on multi-person group activities and simple actions composing those. 


\section{METHOD}
\label{sec:method}
\begin{figure}
\begin{center}
\includegraphics[width=0.38 \textwidth]{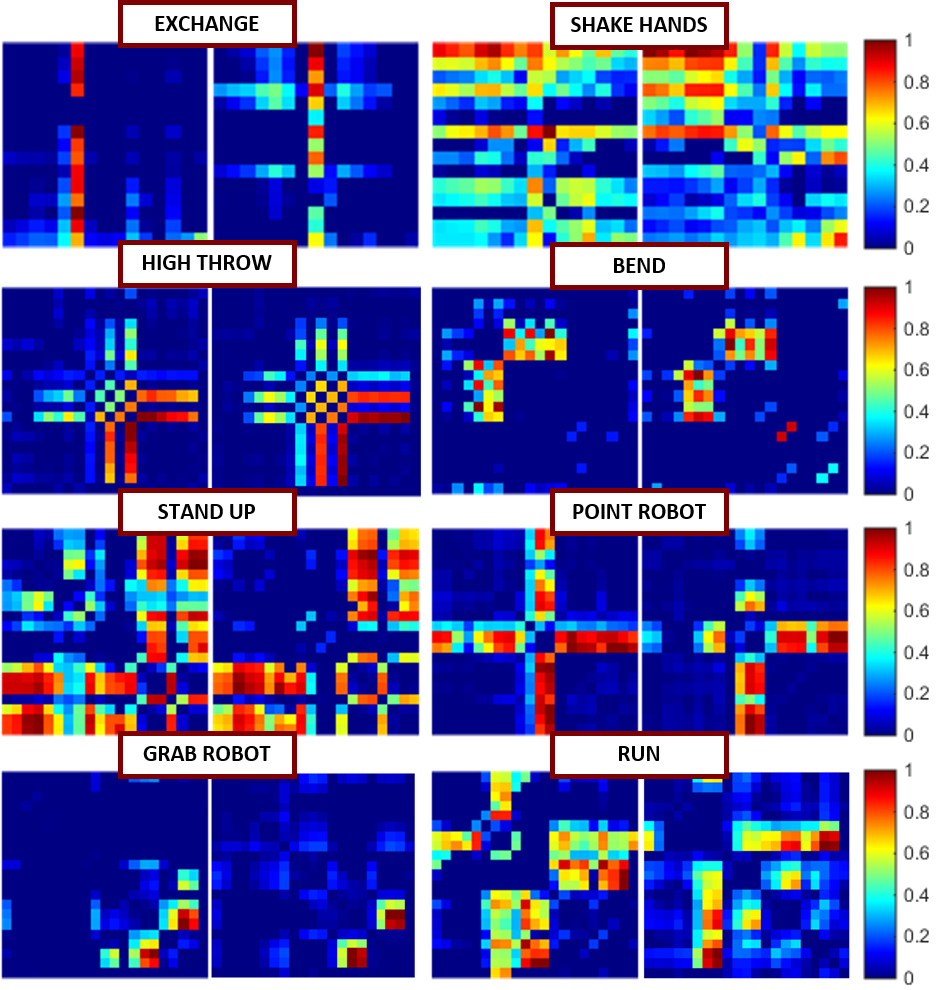}
\end{center}
\caption{Examples of the RHI descriptors on human joint locations. In the first row, RHIs have been extracted from \textit{exchange} and \textit{shake hands} videos, SBU dataset \cite{sbu}. The second row contains examples of \textit{high throw} and \textit{bend} from the MSRAction3D dataset \cite{msr}. In the third row, RHI is extracted from the First-Person RGBD Interaction dataset \cite{lu15}, actions \textit{stand up} and \textit{point robot}. The last row depicts RHIs taken from our dataset. The recorded activities are \textit{grab robot} and \textit{run}.}
\label{fig:RHI}
\end{figure}

We introduce a new \emph{unified} feature descriptor to handle multiple different types of human activities: Relation History Image (RHI). Our descriptor is unified as it is able to describe different types of human activities while maintaining the same format and dimension. Having a unified descriptor is very important in robot-centric scenarios, where we do not know a priori who is interacting with another person, who is by himself, and who is interacting with the robot. In this case, we need classification results on interactions, single person activities and ego-centric interactions to be directly comparable. If we had three distinct classification systems (one for interactions, one for single person actions and one for ego-centric ones), it would not be possible to infer whether an interaction between two persons p1 and p2 is more likely than person p1 sitting and person p2 hugging the robot. Instead, thanks to the unified descriptor, we can train all the activities within the same classification framework and make decisions. At the same time, since the actual matrix is different when extracted from single persons and pairs, it is easy for the classifier to learn differences between single actions and interactions. 

Our approach takes a set of unlabeled RGB-D frames, where multiple people perform activities of different types, and returns a set of pairs with their associated activity classes. We consider a pair formed by the same person selected twice as an acceptable pair to enable single-person action representations. We take advantage of Kinect 2.0 to detect human bodies. Relation History Images (RHIs) from \textbf{each possible pair of appearing persons} is extracted over a sliding window of frames. During training, we know who are the persons that are interacting, who is alone and who is interacting with the robot; the related descriptors model \textit{valid} actions. All the other descriptors extracted from pairs that are not modeling a real activity (non-\textit{valid} actions) are associated to the `null' class. A one-vs-one SVM is trained based on all the RHIs, including those assigned to the `null' class, regardless of their types. When testing, we do not use any annotation on the pairs who are interacting and the pairs who are not. In order to simultaneously recognize the activities and identify the persons performing them, we solve a linear optimization problem, exploiting the SVM responses for each pair in the scene.

For example, in Fig. \ref{fig:dataset}, middle column, the pairs \textit{\{green, blue\}}, \textit{\{pink, yellow\}}, and \textit{\{cyan, cyan\}} are associated to \textit{valid} activities, because the descriptors generated from those pairs are modeling real actions. However, the descriptor extracted from the pair \textit{\{blue,cyan\}} is not describing an interaction between \textit{blue} and \textit{cyan}, because \textit{cyan} is actually on his own, and \textit{blue} is interacting with \textit{green}. Thus, the descriptor extracted from the pair \textit{\{blue,cyan\}} models a non valid action, and it is assigned to a `null' class. Likewise, the pair \textit{\{pink,pink\}} and the pair \textit{\{yellow,yellow\}} will be assigned to the `null' class.

\subsection{Relation History Image}

We propose a new mid-level descriptor for general activity recognition: the Relation History Image (RHI). Given a set of $m$ local regions $(r_1, ..., r_m)$ identified on a person, we extract a local feature vector from each of them $(x_1, ..., x_m)$. We then build a $m \times m$ matrix $\mathbf{R_t}$ for each frame $t$, for each pair in the scene. Each element of $\mathbf{R_t}$ is equal to
\begin{equation}
R_t^{i,j}=K(x_t^i,x_t^j),
\label{eq:kinrhi}
\end{equation}
where $K(\cdot,\cdot)$ is a function that measures the relation between the low-level descriptors extracted from the $i$-th and the $j$-th local regions at time $t$. Notably, $\mathbf{R_t}$ can describe the relation between local regions on two different persons, as well as the relation between local regions on the same person (i.e., RHI can model the relations among one person joints by choosing the same person twice). Two-persons RHIs are showed in Fig. \ref{fig:RHI}, first row. Examples of single-person RHIs, extracted from third-person single actions and human-robot interactions, are showed in Fig. \ref{fig:RHI}, second, third, and fourth rows. Notably, they are symmetric. 

Inspired by the well-known Motion History Image (MHI) \cite{bobick01}, we embed the temporal information in our descriptor computing a series of temporal differences. We consider windows $[t,t+F]$ of $F$ frames and we build a tensor $\mathbf{W_t}$ composed of the differences between all the matrices $\mathbf{R_{t+f}}$, $f \in \{2, ..., F\}$ in the window and the first one:
\begin{equation}
\mathbf{W_t}=\begin{matrix} [\mathbf{R_{t+1}}-\mathbf{R_{t}} & \mathbf{R_{t+2}}-\mathbf{R_{t}} & ... & \mathbf{R_{t+F-1}}-\mathbf{R_{t}}]\end{matrix}.
\end{equation}
We further convolute $\mathbf{W_t}$ with a set of 1D Gaussian filters over the temporal dimension, each of which having $\mu=\dfrac{2t+F}{2}$ and $\sigma^2 \in (0,1)$. In particular, we use three different sigma values: the first one is conceived to consider mostly the difference between the middle frame and the initial frame of the window; the second and the third one linearly increase the weight of the differences between the other frames in the window and the first one. The responses are summed up in the final RHI descriptor:
\begin{equation}
\mathbf{RHI_{t}}=\sum_{j=1}^{s}\mathbf{W_t} \ast h(\mu,\sigma_j^2),
\end{equation}
where $s=3$ is the number of filters employed, $h(\mu,\sigma_j^2)$ is a Gaussian filter applied to the temporal dimension, and $\mathbf{RHI_{t}}$ is the RHI descriptor extracted from the window $[t,t+F]$. Fig. \ref{fig:RHI} depicts some examples of RHI. The only parameter that influences the proposed descriptor is the number of frames on which it is calculated. We experimentally found that a small number of frames, such as $5$ or $8$, performs well in any setting.

\subsubsection{Relation History Image on Joints}
\label{sec:rhijoints}
Joints are the most informative local regions for humans. With the popularity of the Kinect sensor, the possibility of tracking joints automatically has significantly increased, and several works in the literature rely on this information to obtain effective descriptors \cite{wang12,wang132,luo13,vemu14,wu14}. For this reason, we propose a RHI descriptor that represents relations between pairs of joints over time. In this case, relations can be simply the Euclidean distances between pairs of joints:
\begin{equation}
K(\mathbf{x_t^i},\mathbf{x_t^j})=\|\mathbf{x_t^i}-\mathbf{x_t^j}\|,
\end{equation}
where $\mathbf{x_t^i}$ is the 3D position of the $i$-th joint at time $t$ and $\mathbf{x_t^j}$ is the 3D position of the $j$-th joint at time $t$ (see Fig. \ref{fig:rhiextr}). This formulation models variations over time. If two activities are represented by very similar temporal variations but different initial positions (i.e., sitting still or standing still), it is useful to model the initial configuration as well. For this reason, we concatenate to the RHI on joints the mean of the $\mathbf{R_t}$ matrices defined in \ref{eq:kinrhi} over the set of frames, considering a smaller number of joints chosen randomly.

\subsubsection{Relation History Image on Depth}
In order to build a similar structure using depth information, we exploit as low-level feature the so-called $\tau$ test, described in \cite{tau} and used in \cite{stronzo} for activity recognition. Given a depth image patch, the $\tau$ test is computed as follow:
\begin{equation}
\tau(i,j) = \begin{cases} 1, & \mbox{if } d(i)>d(j) \\ 0, & \mbox{otherwise}, \end{cases}
\end{equation}
where $i$ and $j$ represent two pixel locations, and $d(i)$ and $d(j)$ correspond to the depth values of those pixels. We use the modified $\tau$ test proposed in \cite{stronzo}, which adds a second bit to each pixel comparison:
\begin{equation}
\tau_2(i,j) = \begin{cases} 1, & \mbox{if } |d(i)-d(j)|<\epsilon \\ 0, & \mbox{otherwise}. \end{cases}
\end{equation}

We first build a bounding box around each person in the scene, and we split it in $m$ cells, which represent the locally fixed regions. Then, for each cell we sample $P$ pairs of pixels, whose locations are extracted from an isotropic Gaussian distribution:
\begin{equation}
(\mathbf{p_x,p_y}) = i.i.d. N(0, \dfrac{1}{25}\sigma^2),
\end{equation}
where $N(\cdot,\cdot)$ is the Normal distribution. Such locations are kept fixed for the whole dataset, so that they can be coherently compared across different cells and frames. The comparisons between pairs of pixels within a cell are performed in an ordered manner, following the order with which locations have been sampled. From each pair of pixels we obtain a binary value, therefore from $P$ ordered pairs of pixels we retrieve a $P$-long binary string, which describes a cell and constitutes our low-level feature vector. The relation between pairs of cells is represented using the Hamming distance between pairs of strings. Hence, the $K(\cdot,\cdot)$ function used in \ref{eq:kinrhi} in this case is:
\begin{equation}
K(\mathbf{x_t^i},\mathbf{x_t^j})=Hamming(\mathbf{x_t^i},\mathbf{x_t^j}),
\end{equation}
where $\mathbf{x_t^i}$ and $\mathbf{x_t^j}$ represent the binary strings of cell $i$ and cell $j$ at time $t$ (see Fig. \ref{fig:rhiextr}).

\begin{figure}
\begin{center}
\includegraphics[width=0.38 \textwidth]{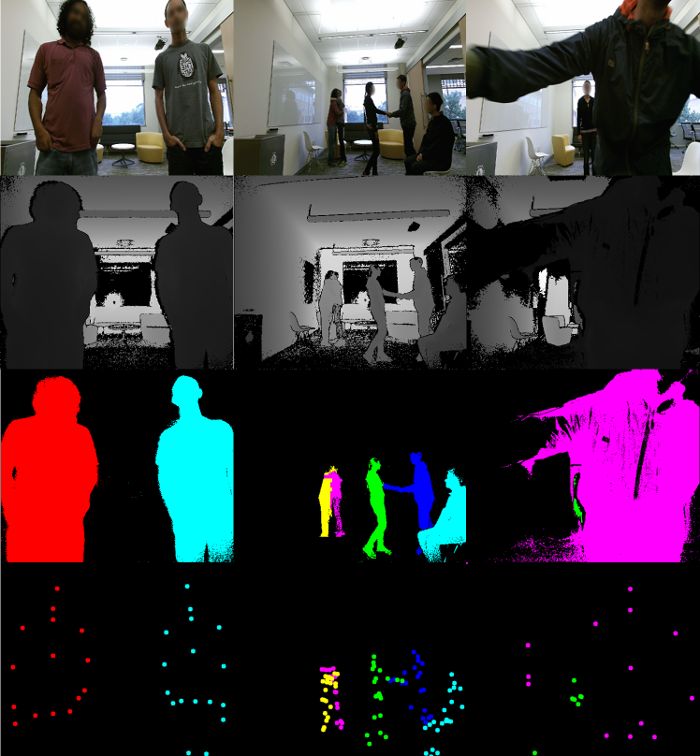}
\end{center}
\caption{Sample images from the new Multy-Type dataset.}
\label{fig:dataset}
\end{figure}

\subsection{Learning and Classification}
\label{sec:learning}

We evaluate temporal \textit{segments}, where a \textit{segment} is defined as a set of frames where each user is performing one activity that lasts the entire time. This means that each pair -- including the pairs formed by the same person selected twice -- can be associated with a single label for the entire segment, including the `null' label, but the activities executed by different pairs can belong to different types. A new segment starts if a new person enters the scene, if someone leaves the scene, or if someone starts performing a different activity. Obviously, a label on a pair can be the same for several sequential segments. For each segment, there is no a priori information on who is interacting with whom, who is on his own and who is interacting with the robot. Therefore, we extract RHIs on skeleton and depth from all the possible pairs. Given a segment where there are $n \geq 2$ persons, we compute a number of descriptors $D(n)$ equal to

\begin{equation}
D(n)=n+\dfrac{n!}{2(n-2)!};
\end{equation}
the first $n$ descriptors are extracted from pairs formed by the same person repeated twice, while the remaining come from all the possible simple combinations without repetitions of the $n$ subjects in pairs. For $n=1$, $D(n)=1$. RHI is computed over a set of $F$ frames, therefore, for each video containing $n$ subjects and $T$ frames, we will extract $D(n)\cdot(T-F)$ RHIs. Then we train a one-vs-one SVM, without distinguishing among different types of activities. Since descriptors are extracted from all the possible combinations of subjects in the scene, some of them do not model valid activities. For this reason, we introduce the `null' class, $c_{k_0}$, which gathers descriptors obtained in such cases. For each test data $RHI_t$ extracted from a set of $F$ frames and from a specific pair, the one-vs-one SVM outputs a set of values $\mathbf{s}$, one for each pair of classes. Each value $s_{k,l}$ is positive if the test data has been classified as belonging to class $k$, negative if belonging to class $l$. Based on these values, given the set $\mathbf{C}$ of the activity classes, we can build a vector of votes $\mathbf{v}=[v_1, v_2, ..., v_{|C|}]$ for each segment, for each pair, where:
\begin{equation}
v_k=\sum_{t=1}^{T}\sum_{l=1, l\neq k}^{|C|} h(RHI_t,c_k,c_l),
\end{equation}
and

\begin{equation}
h(RHI_t,c_k,c_l)=\begin{cases} 1, & \mbox{if } s_{k,l}>0 \\ 0, & \mbox{otherwise}. \end{cases}
\end{equation}
For each segment, we obtain a matrix $\mathbf{V} \in \mathbb{R}^{|C| \times D(n)}$, where element $v_{k,ij}$ indicates the votes obtained from class $c_k$ on the test data extracted from the pair $(u_i,u_j)$. The first $n$ columns contain the votes obtained by descriptors extracted from pairs where the same user is repeated twice, while the remaining columns contain the votes obtained by pairs of different users.

The simultaneous identification and classification is performed using an optimization procedure over the matrix $\mathbf{V}$ so generated: given the votes obtained by all the classes on each possible pair of users, we would like to select the pairs that are likely to perform valid activities (i.e., different from the `null' class), associated to the activity classes that they are performing. The intuition is that the descriptors extracted from pairs that are performing a valid activity $c_k$ should obtain the highest number of votes exactly on that activity. At the same time, the descriptors extracted from pairs that are not performing any valid activity should get the highest votes from the `null' class $c_{k_0}$. At the end of the procedure, each pair in the segment will be associated to one and only one valid label. Therefore, if a pair composed of the same user $(u_i,u_i)$ is assigned to a valid label, all the pairs formed with $u_i$ will be assigned to the `null' class $c_{k_0}$. Likewise, if a pair $(u_i,u_j)$ is assigned to a valid class, then the pairs $(u_i,u_i)$ and $(u_j,u_j)$ will be labeled as $c_{k_0}$. Given the set of users $\mathbf{U}$, the set of activities $\mathbf{C}$ and the matrix of votes $\mathbf{V}$ for one segment, we formulate an assignment problem slightly modified. We create a set of variables $\Phi$, one for each element of the matrix $\mathbf{V}$. Then, we solve:
\begin{equation}
\begin{aligned}
\Phi^* = & \underset{\Phi}{\text{argmax}}
& & \sum_{k=1}^{|C|}\sum_{i=1}^n\sum_{j=1}^n v_{k,ij}\phi_{k,ij} \\
& \text{s. t.}
& & \sum_{k=1}^{|C|}\phi_{k,ij}=1; \forall u_i,u_j; \\
& \text{}
& & \sum_{k=1,k\neq k_0}^{|C|}\phi_{k,ii}+\sum_{k=1,k\neq k_0}^{|C|}\sum_{j=1,j\neq i}^{n}\phi_{k,ij}=1; \forall u_i; \\
& \text{}
& & \phi_{k,ij} \in \{0,1\}; \forall u_i,u_j,c_k. \\
\end{aligned}
\end{equation}
The objective function represents the fact that we want to choose classes and users that maximize the votes obtained from the classifier. The first constraint models the fact that each pair has to be assigned to only one class. The second constraint handles the assignment between pairs composed of the same user and pairs of different users. We solve this problem using a branch and cut methodology implemented within the \textit{cplex} IBM library.

\begin{figure}
\begin{center}
\includegraphics[width=0.38 \textwidth]{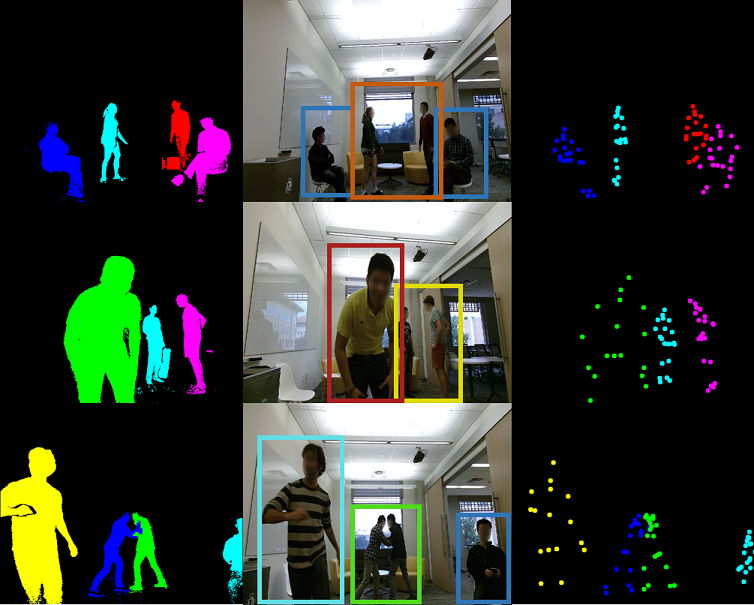}
\end{center}
\caption{Some labeled examples from our dataset. First row: the orange couple is approaching, while the two persons on the sides are sitting. Second row: the person in the red box is talking to the robot, while the two persons behind him are talking. Third row: one person is sitting, one running and two pushing.}
\label{fig:seg}
\end{figure}

\section{EXPERIMENTS}
\label{sec:experiments}
This section presents our experimental results on multiple datasets. RHI performance is compared to other state-of-the-art descriptors on our new dataset as well as on three public video datasets. 

\begin{table*}
\caption{The table shows results achieved by our method and other state-of-the-art algorithms on publicly available datasets and on our new dataset.}
\small
\center
\begin{tabular}{|c|c|c|c|c|c|c|c|}
\hline
\bf{Method} & \bf{SBU} & \bf{FP} & \bf{MSR} & \bf{MT-Acc} & \bf{MT-F1} & \bf{MT-F2} & \bf{Real-time} \\ \hline \hline
Yun et al.~\cite{sbu} & 80.03\% & - & - & - & - & - & Y  \\ \hline
Ji et al.~\cite{ji14} & 86.9\% & - & - & - & - & - & Y  \\ \hline
Oreifej et al.~\cite{hon4d} & 77.0\% & 45.55\% & 88.36\% & 68.07\% & 0.7315 & 0.7870 & \textcolor{red}{N} \\ \hline
Xia et al.~\cite{lu13} & 42.69\% & 53.25\% & 37.76\% & 28.38\% & 0.3891 & 0.46 & \textcolor{red}{N} \\ \hline
Laptev et al.~\cite{laptev03} & 66.28\% & 50.84\% & - & 38.59\% & 0.4519 & 0.5641 & \textcolor{red}{N} \\ \hline
Xia et al.~\cite{luw13} & 41.88\% & 70.0\% & 78.97\% & 47.06\% & 0.4434 & 0.5075 & Y  \\ \hline
Xia et al.~\cite{lu15} & - & 83.7\% & - & - & - & - & \textcolor{red}{N} \\ \hline
Li et al.~\cite{msr} & - & - & 74.7\% & - & - & - & ? \\ \hline
Wang et al.~\cite{wang122} & - & - & 86.5\% & - & - & - & ? \\ \hline
Wang et al.~\cite{wang12} & - & - & 88.2\% & - & - & - & ? \\ \hline
Wang et al.~\cite{wang132} & - & - & 90.22\% & - & - & - & ?  \\ \hline
Evangelidis et al.~\cite{ev14} & - & - & 89.86\% & - & - & - & ? \\ \hline
Chaaraoui et al.~\cite{cha13} & - & - & 91.8\% & - & - & - & Y  \\ \hline
Vemulapalli et al.~\cite{vemu14} & - & - & 92.46\% & - & - & - & ? \\ \hline
Luo et al.~\cite{luo13} & - & - & \bf{96.70\%} & - & - & - & ?  \\ \hline
\bf{RHI (ours)} & \bf{93.08\%} & \bf{85.94\%} & 95.38\% & \bf{76.67\%} & \bf{0.7954} & \bf{0.8633} & Y \\ \hline
\end{tabular}
\label{tab:final}
\end{table*}

\subsection{Experiments on our Multi-Type Dataset}
\label{sec:dataset}
\subsubsection{Dataset}
\begin{figure}
\begin{center}
\includegraphics[width=0.36 \textwidth]{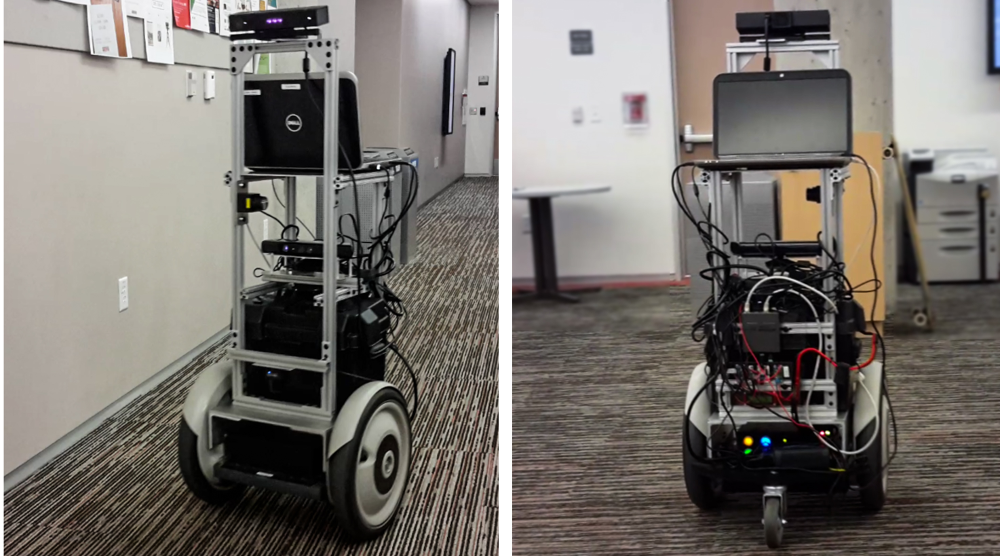}
\end{center}
\caption{This figure shows the robot that we have used for our experiments. The Kinect v2.0 is mounted on its head.}
\label{fig:rob}
\end{figure}

We collected a new RGB-D dataset, called Multi-Type Dataset, which includes videos where multiple types of activities are performed concurrently and sequentially. We took advantage of Kinect 2.0 for the tracking. The sensor has been mounted on an autonomous non-humanoid robot (see Fig. \ref{fig:rob}), which is designed to move around in a building populated by students. We recreated a natural environment, where students meet, wait to go to class, or interact with the robot. We asked $20$ participants divided in $5$ groups of $4-5$ persons to perform $12$ different sequences of activities. Each sequence is characterized by the presence of $2$ to $5$ persons performing actions, with different body orientations and at different scales. An example of a sequence could be: a pair of persons get close, shake hands and start talking while a person sits by himself, and another person approaches the robot and clears the path to avoid it. We asked the subjects to act naturally and with their own timing, therefore the sequences are always different. We defined $16$ basic activities: $6$ two-person interactions, \textit{approach, hug, shake hands, push, talk, wave}, $6$ first-person interactions, \textit{approach robot, wave to the robot, point the robot, clear the path, talk to the robot, grab the robot}, and $4$ single activities, \textit{sit, stand up, walk, run}.

We collected RGB, depth and skeletal data (see Fig. \ref{fig:dataset}). The images are recorded at around $20$ fps. The depth images are $512 \times 424$ 16-bits single channel. RGB images are $640 \times 480$ three-channels. The skeleton is composed of $25$ joints, whose positions are provided in both 3D and 2D. Some images extracted from the dataset are shown in Fig. \ref{fig:dataset}, while Fig. \ref{fig:seg} depicts some labeled examples.

\subsubsection{Experimental Setting}
Given multiple continuous videos containing a total of 288 activity executions which may temporally overlap, 523 samples are extracted including action samples as well as non-valid action samples (i.e., `null' class samples not corresponding to any activity) for the training/testing of our approach and baselines. The one-vs-one SVM is trained based on all these samples by using corresponding action samples as positive samples and all the others as negative samples. We performed a leave-one-group-out cross-validation treating $4$ groups of people as training set and $1$ group as test set.

For the testing, we carried out two experiments. In the first one, we assume that there is no information on who are the pairs that perform valid actions. As described in Sec. \ref{sec:learning}, a label is associated to each possible pair in the scene. To evaluate the performance of our method, we use precision and recall in the following fashion: we treat detections as true positives if the pairs that have been detected as performing a valid activity and, at the same time, the classified activity are correct; the pairs correctly identified as performing a valid activity, but whose activity has been classified incorrectly, are treated as false positives; false negatives correspond to pairs that have been incorrectly identified as belonging to the `null' class. In the second experiment, we assume that the pairs who are performing valid activities are known. In this case, we evaluate the effectiveness of descriptors in terms of classification accuracy.

\subsubsection{Experimental Results and Comparison}
In order to establish a baseline, HON4D \cite{hon4d}, STIP \cite{laptev03} and DSTIP \cite{lu13} are assessed on our dataset. Being appearance-based descriptors, they can be adapted to handle multiple activity types just by cropping the image around a pair, and by using the crop to extract the feature vectors associated to that pair. For HON4D and DSTIP, whose source code is available, we tested several sets of parameters and we report here the highest results obtained. 

Table \ref{tab:final}, column \textbf{MT-F1} and \textbf{MT-F2}, summarizes the comparison between our descriptor and HON4D, STIP and DSTIP. The recognition accuracy, computed assuming that the true pairs are known, is reported in Table \ref{tab:final}, column \textbf{MT-Acc}. To take into account the randomness in the dictionary generation step, we report the mean value over $10$ different trials for STIP and DSTIP. As the table shows, RHI outperforms any other general purpose feature descriptor. This is probably due to the fact that RHI can model situations where the body configuration is very important.

To show that RHI is more flexible than other joint-based methods, which are usually not unified, we extend HOJ3D \cite{luw13} so that it becomes applicable to our scenario: when considering pairs constituted by different persons, we concatenated the two HOJ3D; when considering a pair composed of the same subject repeated twice, we concatenated a set of zeros at the end of the HOJ3D computed on the subject. We test this simple extension on the SBU dataset too, which has been recorded to recognize interactions. Table \ref{tab:final} shows that the results achieved by this method on our dataset, as well as on the SBU dataset, are poor. This confirms that it is not trivial to find a way to extend conventional non-unified descriptors to recognize multiple types of activities.

\subsection{Comparison on Public Datasets}
We assess our new RHI descriptor on three publicly available datasets. In this case, we only need to evaluate its recognition accuracy. We compute the RHI descriptor on each set of $F$ frames in every video, therefore, if the video is composed of $T$ frames, we extract $(T-F)$ RHI descriptors, which constitute positive examples for the classifiers. We proceed in training non-linear SVMs with Radial Basis kernel in a one-versus-all fashion. The kernel parameters are estimated during cross-validation. Given a test example, the SVM scores are calculated for each of the $m$ classifiers and for each RHI. The test sample is labeled as belonging to the class that obtained the highest score:

\begin{equation}
c=\argmax_{i=1, \ldots, m} \sum_{j=1}^{T}h_i(RHI_{j,j+k}),
\end{equation}
where $h_i(\cdot)$ is the SVM $i$-th classifier. For the following experiments, we found that computing the RHI on windows of $5$ frames provides the best accuracy.

\subsubsection{Two-Person Interactions; SBU Dataset}
\label{sec:sbu}
The SBU Kinect Interaction Dataset \cite{sbu} has been recorded for interaction recognition purposes. It contains two person interactions videos from a third-person perspective. This is the first RGBD dataset for two person interactions, and the only one containing human joint information. The dataset contains examples of eight different interactions: \textit{approaching, departing, kicking, punching, pushing, hugging, shaking hands and exchanging something}. We perform a 5-fold cross validation following the instructions provided in \cite{sbu}. Table \ref{tab:final} shows the results compared to the state-of-the-art in the column called \textbf{SBU}. As Table \ref{tab:final} reports, our method outperforms previous works on this dataset.

\subsubsection{First-person Interactions; First-Person RGBD Interaction Dataset}
\label{sec:ut}
The non-humanoid robot dataset presented in \cite{lu15} has been collected to recognize first-person interactions. In this dataset, people only interact with (or react to) the person/robot who wears the camera, thus it is significantly different from the dataset considered in Sec. \ref{sec:sbu}. We follow the instructions provided by \cite{lu15} to generate the final accuracy. Our results with respect to state-of-the-art approaches are shown in Table \ref{tab:final}, column \textbf{FP}. As Table \ref{tab:final} reports, we achieve the highest accuracy.

\subsubsection{Third-person Actions; MSRAction3D Dataset}
\label{sec:msr}
The MSRAction3D dataset \cite{msr} contains video where a single person performs simple actions, thus is again different from the previously considered datasets. It provides skeleton and depth information of $20$ different actions. In this case we use, as relational information on the joint locations, the same displacement vector that has been used in \cite{wang12}. This procedure provides the opportunity to evaluate the actual effect of the temporal structure in our RHI. We follow the instructions provided in \cite{msr} to compute the final accuracy using the cross-subject test, which is the most challenging. In particular, we divide the dataset in three subsets (AS1, AS2 and AS3), and we compute the accuracy on them separately using $5$ subjects for training and $5$ subjects for testing. We compare our results with several state-of-the-art methods in Table \ref{tab:final}, column \textbf{MSR}. We only listed the works that used the cross-subject test, as indicated in \cite{msr}. Notably, RHI outperforms the method in \cite{wang12}; this shows that the temporal structure plays an important role to improve the effectiveness of joint-based descriptors.

\subsection{Discussions}
Table \ref{tab:final} shows that RHI is not only the most effective descriptor on our new dataset, but also achieves excellent performance on all the datasets we have considered. The methods that report a `-' in the Table have not been tested on the related datasets. Even though some of these approaches obtained reasonable results on classifying activities of a single type, they were never tested for realistic scenarios where activities of multiple different types are present at the same time, and did not consider learning representations for such situations at all. Such methods do not have an ability to directly compare activities of different types, and thus are not suitable for classifying/detecting them. The only descriptors that could be more easily extended to our setting are appearance-based descriptors such as HON4D, STIP and DSTIP. However, as the experiments confirm, they are not sufficient to model the scenario we have proposed. 

\subsubsection{Computation time}
In Table \ref{tab:final}, we also report the ability of the descriptors to be extracted in real-time. In our case, each RHI computation takes less than $1$ msec on the joints and less than $20$ msec on the depth on a common laptop. The final computational cost is quadratic with respect to the number of persons in the scene. Note that all these can still be done in real-time with a support from modern CPU/GPU architectures. In our approach, each RHI is independent from the others. This means that we are able to parallelize all these RHI computations using modern GPUs that handle a large number of threads simultaneously. 


\subsubsection{Occlusion}
We also want to mention that our dataset contains multiple types of partial occlusions, more than most of the previous RGBD activity datasets. In order to confirm this, we measured the `occlusion level' of each dataset including ours: we computed the largest occlusion between any two persons present in each video and averaged such values over all the videos. More specifically, we considered bounding boxes of every pair of appearing persons, and computed their intersection divided by the size of the larger bounding box. Our dataset presents the occlusion level of 0.78. The SBU dataset \cite{sbu} has the occlusion of 0.44, while the other two public datasets \cite{msr,lu15} have 0 occlusion (i.e., they are datasets with only one person in each video).



\section{CONCLUSION}
\label{sec:conclusion}
Our work addresses the problem of labeling a complex robot-centric scene where multiple persons perform different activity types concurrently and sequentially. Toward this goal, we proposed RHI, a new unified descriptor that does not depend on any sensitive parameter and can be computed in real-time. We further propose a new optimization-based method that enables the simultaneous classification of activities and identification of the subjects involved. Experimental results confirm that RHI outperforms previous works on publicly available datasets as well as on the newly collected Multi-Type Dataset, a new RGB-D dataset which can be useful to the community for future benchmarking. Few traditional descriptors can handle a scenario where different types of activities are performed at the same time. We show that, even those that can be assessed on our dataset, obtain poor performances with respect to ours. 

At this stage, our approach does not handle group activities with three or more persons: if three persons are doing group-conversation, two persons will be labeled as doing "talking" interaction and the single action with the highest confidence will be assigned to the remaining person. A possible future direction is to handle group activities using graph properties in the optimization procedure, finding persons with the same activity. Another possible extension regards improving the optimization procedure, by substituting the SVM-based voting strategy with methods which output a confidence measure along with label predictions. In this paper, we focused on developing the unified descriptor, RHI, and further improving the optimization remains a future challenge.


\addtolength{\textheight}{-12cm}   

{\flushleft\textbf{Acknowledgement.} This research was supported in part by the Army Research Laboratory under Cooperative Agreement Number W911NF-10-2-0016.}



{\small
\bibliographystyle{IEEEtran}
\bibliography{IEEEabrv,egbib}
}

\end{document}